\documentclass[11pt,a4paper]{article}
\usepackage[hyperref]{acl2018}
\usepackage{algorithm, algpseudocode}
\usepackage{times}
\usepackage{comment}
\usepackage{latexsym}
\usepackage{multirow}
\usepackage{amsmath}
\usepackage{CJK}

\usepackage{url}

\aclfinalcopy 
\def\aclpaperid{1365} 

\newcommand*{\affaddr}[1]{#1} 
\newcommand*{\affmark}[1][*]{\textsuperscript{#1}}
\newcommand*{\email}[1]{\texttt{#1}}

\title{Learning Domain-Sensitive and Sentiment-Aware Word Embeddings\thanks{\ \ This work was partially done when Bei Shi was an intern at Tencent AI Lab. This project is substantially supported by a grant from the Research Grant Council of the Hong Kong Special Administrative Region, China (Project Code: 14203414).}}

\author{
Bei Shi\affmark[1], Zihao Fu\affmark[1], Lidong Bing\affmark[2] and
Wai Lam\affmark[1]\\
\affaddr{\affmark[1]Department of Systems Engineering and Engineering Management\\
The Chinese University of Hong Kong, Hong Kong}\\
\affaddr{\affmark[2]Tencent AI Lab, Shenzhen, China}\\
\email{\{bshi,zhfu,wlam\}@se.cuhk.edu.hk}\\
\email{lyndonbing@tencent.com}\\
}

\date{}

\begin{document}
\maketitle
\begin{abstract}
Word embeddings have been widely used in sentiment classification because of their efficacy for semantic representations of words. Given reviews from different domains, some existing methods for word embeddings exploit sentiment information, but they cannot produce domain-sensitive embeddings. On the other hand, some other existing methods can generate domain-sensitive word embeddings, but they cannot distinguish words with similar contexts but opposite sentiment polarity. We propose a new method for learning domain-sensitive and sentiment-aware embeddings that simultaneously capture the information of sentiment semantics and domain sensitivity of individual words. Our method can automatically determine and produce domain-common embeddings and domain-specific embeddings. The differentiation of domain-common and domain-specific words enables the advantage of data augmentation of common semantics from multiple domains and capture the varied semantics of specific words from different domains at the same time. Experimental results show that our model provides an effective way to learn domain-sensitive and sentiment-aware word embeddings which benefit sentiment classification at both sentence level and lexicon term level.
\end{abstract}

\section{Introduction}
Sentiment classification aims to predict the sentiment polarity, such as ``positive'' or ``negative'', over a piece of review. It has been a long-standing research topic because of its importance for many applications such as social media analysis, e-commerce, and marketing~\cite{liu2012sentiment, pang2008opinion}. Deep learning has brought in progress in various NLP tasks, including sentiment classification. Some researchers focus on designing RNN or CNN based models for predicting sentence level~\cite{kim2014convolutional} or aspect level sentiment~\cite{li2018transformation, chen2017recurrent, wang2016attention}. These works directly take the word embeddings pre-trained for general purpose as initial word representations and may conduct fine tuning in the training process. Some other researchers look into the problem of learning task-specific word embeddings for sentiment classification aiming at solving some limitations of applying general pre-trained word embeddings. For example, Tang et al.~\shortcite{tang2014learning} develop a neural network model to convey sentiment information in the word embeddings. As a result, the learned embeddings are \textbf{sentiment-aware} and able to distinguish words with similar syntactic context but opposite sentiment polarity, such as the words ``good'' and ``bad''. In fact, sentiment information can be easily obtained or derived in large scale from some data sources (e.g., the ratings provided by users), which allows reliable learning of such sentiment-aware embeddings.

Apart from these words (e.g. ``good'' and ``bad'') with consistent sentiment polarity in different contexts, the polarity of some sentiment words is \textbf{domain-sensitive}. For example, the word ``lightweight'' usually connotes a positive sentiment in the electronics domain since a lightweight device is easier to carry. In contrast, in the movie domain, the word ``lightweight'' usually connotes a negative opinion describing movies that do not invoke deep thoughts among the audience. This observation motivates the study of learning domain-sensitive word representations~\cite{yang2017simple,bollegala2015unsupervised,bollegala2014learning}. They basically learn separate embeddings of the same word for different domains. To bridge the semantics of individual embedding spaces, they select a subset of words that are likely to be domain-insensitive and align the dimensions of their embeddings. However, the sentiment information is not exploited in these methods although they intend to tackle the task of sentiment classification.

In this paper, we aim at learning word embeddings that are both domain-sensitive and sentiment-aware. Our proposed method can jointly model the sentiment semantics and domain specificity of words, expecting the learned embeddings to achieve superior performance for the task of sentiment classification. Specifically, our method can automatically determine and produce domain-common embeddings and domain-specific embeddings. Domain-common embeddings represent the fact that the semantics of a word including its sentiment and meaning in different domains are very similar. For example, the words ``good'' and ``interesting'' are usually domain-common and convey consistent semantic meanings and positive sentiments in different domains. Thus, they should have similar embeddings across domains. On the other hand, domain-specific word embeddings represent the fact that the sentiments or meanings across domains are different. For example, the word ``lightweight'' represents different sentiment polarities in the electronics domain and the movie domain. Moreover, some polysemous words have different meanings in different domains. For example, the term ``apple'' refers to the famous technology company in the electronics domain or a kind of fruit in the food domain. 

Our model exploits the information of sentiment labels and context words to distinguish domain-common and domain-specific words. If a word has similar sentiments and contexts across domains, it indicates that the word has common semantics in these domains, and thus it is treated as domain-common. Otherwise, the word is considered as domain-specific. The learning of domain-common embeddings can allow the advantage of data augmentation of common semantics of multiple domains, and meanwhile, domain-specific embeddings allow us to capture the varied semantics of specific words in different domains. Specifically, for each word in the vocabulary, we design a distribution to depict the probability of the word being domain-common. The inference of the probability distribution is conducted based on the observed sentiments and contexts. As mentioned above, we also exploit the information of sentiment labels for the learning of word embeddings that can distinguish words with similar syntactic context but opposite sentiment polarity.

To demonstrate the advantages of our domain-sensitive and sentiment-aware word embeddings, we conduct experiments on four domains, including books, DVSs, electronics, and kitchen appliances. The experimental results show that our model can outperform the state-of-the-art models on the task of sentence level sentiment classification. Moreover, we conduct lexicon term sentiment classification in two common sentiment lexicon sets to evaluate the effectiveness of our sentiment-aware embeddings learned from multiple domains, and it shows that our model outperforms the state-of-the-art models on most domains.

\section{Related Works}
Traditional vector space models encode individual words using the one-hot representation, namely, a high-dimensional vector with all zeroes except in one component corresponding to that word~\cite{baeza1999modern}. Such representations suffer from the curse of dimensionality, as there are many components in these vectors due to the vocabulary size. Another drawback is that semantic relatedness of words cannot be modeled using such representations. To address these shortcomings, Rumelhart et al.~\shortcite{rumelhart1988learning} propose to use distributed word representation instead, called word embeddings. Several techniques for generating such representations have been investigated. For example, Bengio et al. propose a neural network architecture for this purpose~\cite{bengio2003neural,bengio2009learning}. Later, Mikolov et al.~\shortcite{mikolov2013distributed} propose two methods that are considerably more efficient, namely skip-gram and CBOW. This work has made it possible to learn word embeddings from large data sets, which has led to the current popularity of word embeddings. Word embedding models have been applied to many tasks, such as named entity recognition~\cite{turian2010word}, word sense disambiguation~\cite{collobert2011natural, iacobacci2016embeddings, Zhang.Hasan.3, vachik_zhang_arxiv}, parsing~\cite{roth2016neural}, and document classification~\cite{tang2014building, tang2014learning, shi2017jointly}.

Sentiment classification has been a long-standing research topic~\cite{liu2012sentiment, pang2008opinion, chen2017recurrent,moraes2013document}. Given a review, the task aims at predicting the sentiment polarity on the sentence level~\cite{kim2014convolutional} or the aspect level~\cite{li2018transformation, chen2017recurrent}. Supervised learning algorithms have been widely used in sentiment classification~\cite{pang2002thumbs}. People usually use different expressions of sentiment semantics in different domains. Due to the mismatch between domain-specific words, a sentiment classifier trained in one domain may not work well when it is directly applied to other domains. Thus cross-domain sentiment classification algorithms have been explored~\cite{pan2010cross,li2009knowledge,glorot2011domain}. These works usually find common feature spaces across domains and then share learned parameters from the source domain to the target domain. For example, Pan et al.~\shortcite{pan2010cross} propose a spectral feature alignment algorithm to align words from different domains into unified clusters. Then the clusters can be used to reduce the gap between words of the two domains, which can be used to train sentiment classifiers in the target domain. Compared with the above works, our model focuses on learning both domain-common and domain-specific embeddings given reviews from all the domains instead of only transferring the common semantics from the source domain to the target domain.

Some researchers have proposed some methods to learn task-specific word embeddings for sentiment classification~\cite{tang2014building,tang2014learning}. Tang et al.~\shortcite{tang2014learning} propose a model named SSWE to learn sentiment-aware embedding via incorporating sentiment polarity of texts in the loss functions of neural networks. Without the consideration of varied semantics of domain-specific words in different domains, their model cannot learn sentiment-aware embeddings across multiple domains. Some works have been proposed to learn word representations considering multiple domains~\cite{yang2017simple,bach2016cross,bollegala2015unsupervised}. Most of them learn separate embeddings of the same word for different domains. Then they choose pivot words according to frequency-based statistical measures to bridge the semantics of individual embedding spaces. A regularization formulation enforcing that word representations of pivot words should be similar in different domains is added into the original word embedding framework. For example, Yang et al.~\shortcite{yang2017simple} use S{\o}rensen-Dice coefficient~\cite{sorensen1948method} for detecting pivot words and learn word representations across domains. Even though they evaluate the model via the task of sentiment classification, sentiment information associated with the reviews are not considered in the learned embeddings. Moreover, the selection of pivot words is according to frequency-based statistical measures in the above works. In our model, the domain-common words are jointly determined by sentiment information and context words. 

\section{Model Description}
We propose a new model, named DSE, for learning \textbf{D}omain-sensitive and \textbf{S}entiment-aware word \textbf{E}mbeddings. For presentation clarity, we describe DSE based on two domains. Note that it can be easily extended for more than two domains, and we remark on how to extend near the end of this section.

\subsection{Design of Embeddings}
We assume that the input consists of text reviews of two domains, namely $\mathcal{D}^p$ and $\mathcal{D}^q$. Each review $r$ in $\mathcal{D}^p$ and $\mathcal{D}^q$ is associated with a sentiment label $y$ which can take on the value of $1$ and $0$ denoting that the sentiment of the review is positive and negative respectively.

In our DSE model, each word $w$ in the whole vocabulary $\Lambda$ is associated with a domain-common vector $U_w^c$ and two domain-specific vectors, namely $U_w^p$ specific to the domain $p$ and $U_w^q$ specific to the domain $q$. The dimension of these vectors is $d$. The design of $U_w^c$, $U_w^p$ and $U_w^q$  reflects one characteristic of our model: allowing a word to have different semantics across different domains.  The semantic of each word includes not only the semantic meaning but also the sentiment orientation of the word. If the semantic of $w$ is consistent in the domains $p$ and $q$, we use the vector $U_w^c$ for both domains. Otherwise, $w$ is represented by $U_w^p$ and $U_w^q$ for $p$ and $q$ respectively.

In traditional cross-domain word embedding methods~\cite{yang2017simple, bollegala2015unsupervised, bollegala2016cross}, each word is represented by different vectors in different domains without differentiation of domain-common and domain-specific words. In contrast to these methods, for each word $w$, we use a latent variable $z_w$ to depict its domain commonality. When $z_w = 1$, it means that $w$ is common in both domains. Otherwise, $w$ is specific to the domain $p$ or the domain $q$.

In the standard skip-gram model~\cite{mikolov2013distributed}, the probability of predicting the context words is only affected by the relatedness with the target words. In our DSE model, predicting the context words also depends on the domain-commonality of the target word, i.e $z_w$. For example, assume that there are two domains, e.g. the electronics
domain and the movie domain. If $z_w = 1$, it indicates a high probability of generating some domain-common words such as ``good'', ``bad'' or ``satisfied''. Otherwise, the domain-specific words are more likely to be generated such as ``reliable'', ``cheap'' or ``compacts'' for the electronics domain. For a word $w$, we assume that the probability of predicting the context word $w_t$ is formulated as follows:
\begin{equation}
  \label{eq:context}
  p(w_t | w) = \sum_{k \in{\{0, 1\}}}p(w_t|w, z_w=k)p(z_w=k)
\end{equation}

If $w$ is a domain-common word without differentiating $p$ and $q$, the probability of predicting $w_t$ can be defined as:
\begin{equation}
  \label{eq:w_t}
  p(w_t|w, z_w=1)=\frac{\exp(U_w^c \cdot V_{w_t})}{\sum_{w^{\prime}\in\Lambda}\exp(U_w^c \cdot V_{w^{\prime}})}
  \end{equation}
where $\Lambda$ is the whole vocabulary and $V_{w^{\prime}}$ is the output vector of the word $w^{\prime}$.

If $w$ is a domain-specific word, the probability of $p(w_t | w, z_w=0)$ is specific to the occurrence of $w$ in $\mathcal{D}^p$ or $\mathcal{D}^q$. For individual training instances, the occurrences of $w$ in $\mathcal{D}^p$ or $\mathcal{D}^q$ have been established. Then the probability of $p(w_t|w, z_w=0)$ can be defined as follows:
\begin{equation}
  \label{eq:w_t_z}
  \begin{split}
  p(w_t|w, z_w=0)=
  \begin{cases}
    \frac{\exp(U_w^p \cdot
      V_{w_t})}{\sum_{w^{\prime}\in\Lambda}\exp(U_w^p \cdot
      V_{w^{\prime}})}, \text{if } w \in \mathcal{D}^p \\
    \\
    \frac{\exp(U_w^q \cdot
      V_{w_t})}{\sum_{w^{\prime}\in\Lambda}\exp(U_w^q \cdot
      V_{w^{\prime}})}, \text{if } w \in \mathcal{D}^q \\
  \end{cases}
  \end{split}
  \end{equation}
  
\subsection{Exploiting Sentiment Information}

In our DSE model, the prediction of review sentiment depends on not only the text information but also the domain-commonality. For example, the domain-common word ``good'' has high probability to be positive in different reviews across multiple domains. However, for the word ``lightweight'', it would be positive in the electronics domain, but negative in the movie domain. We define the polarity $y_w$ of each word $w$ to be consistent with the sentiment label of the review: if we observe that a review is associated with a positive label, the words in the review are associated with a positive label too. Then, the probability of predicting the sentiment for the word $w$ can be defined as:
 \begin{equation}
   \label{eq:senti}
   p(y_w | w) = \sum_{k \in{\{0, 1\}}}p(y_w|w, z_w=k)p(z_w=k)
 \end{equation}
 If $z_w = 1$, the word $w$ is a domain-common word. The probability $p(y_w = 1|w, z_w=1)$ can be defined as:
 \begin{equation}
   p(y_w = 1|w, z_w=1) = \sigma(U_w^c \cdot \mathbf{s})
 \end{equation}
 where $\sigma(\cdot)$ is the sigmoid function and the vector $\mathbf{s}$ with dimension $d$ represents the boundary of the sentiment. Moreover, we have:
 \begin{equation}
   p(y_w = 0|w, z_w=1) = 1 - p(y_w=1|w, z_w=1)
   \end{equation}

  If $w$ is a domain-specific word, similarly, the probability $p(y_w=1 | w, z_w = 0)$ is defined as:
  \begin{equation}
    \begin{split}
    p(y_w=1|w, z_w=0) =
    \begin{cases}
       \sigma(U_w^p \cdot \mathbf{s}) & \text{if } w \in
      \mathcal{D}^p \\
       \sigma(U_w^q \cdot \mathbf{s}) & \text{if } w \in \mathcal{D}^q \\
    \end{cases}
    \end{split}
  \end{equation}

  \subsection{Inference Algorithm}

\begin{algorithm}[t]
  \caption{EM negative sampling for DSE
}
  \label{alg:em}
  \begin{algorithmic}[1]
    \State Initialize $U_w^c$, $U_w^p$, $U_w^q$, $V$, $\mathbf{s}$, $p(z_w)$
    \For {$\textit{iter} = 1$ to $\textit{Max}\_\textit{iter}$}
    \For{each review $r$ in $\mathcal{D}^p$ and $\mathcal{D}^q$}
    \For{each word $w$ in $r$}
    \State Sample negative instances from the distribution P.
    \State Update $p(z_w|w, c_w, y_w)$ by Eq.~\ref{eq:post} and Eq.~\ref{eq:sim} respectively.
    \EndFor
    \EndFor
    \State Update $p(z_w)$ using Eq.~\ref{eq:z_w}
    \State Update $U_w^c$, $U_w^p$, $U_w^q$, $V$, $\mathbf{s}$ via Maximizing Eq.~\ref{eq:qu}
   \EndFor
   \end{algorithmic}
  \end{algorithm}
  
 We need an inference method that can learn, given $\mathcal{D}^p$ and $\mathcal{D}^q$, the values of the model parameters, namely, the domain-common embedding $U_w^c$, and the domain-specific embeddings $U_w^p$ and $U_w^q$, as well as the domain-commonality distribution $p(z_w)$ for each word $w$. Our inference method combines the Expectation-Maximization (EM) method with a negative sampling scheme. It is summarized in Algorithm~\ref{alg:em}. In the E-step, we use the Bayes rule to evaluate the posterior distribution of $z_w$ for each word and derive the objective function. In the M-step, we maximize the objective function with the gradient descent method and update the corresponding embeddings $U_w^c$, $U_w^p$ and $U_w^q$.

With the input of $\mathcal{D}^p$ and $\mathcal{D}^q$, the likelihood function of the whole training set is:
  \begin{equation}
    \label{eq:obj}
    \mathcal{L} = \mathcal{L}^p + \mathcal{L}^q
  \end{equation}
  where $\mathcal{L}^p$ and $\mathcal{L}^q$ are the likelihood of $\mathcal{D}^p$ and $\mathcal{D}^q$ respectively.
 
  For each review $r$ from $\mathcal{D}^p$, to learn domain-specific and sentiment-aware embeddings, we wish to predict the sentiment label and context words together. Therefore, the likelihood function is defined as follows:
  \begin{equation}
    \mathcal{L}^p = \sum_{r \in \mathcal{D}^p}\sum_{w \in r}\log p(y_w, c_w|w)
    \end{equation}
  where $y_w$ is the sentiment label and $c_w$ is the set of context words of $w$. For the simplification of the model, we assume that the sentiment label $y_w$ and the context words $c_w$ of the word $w$ are conditionally dependent. Then the likelihood $\mathcal{L}^p$ can be rewritten as:
  \begin{equation}
    \begin{split}
    \mathcal{L}^p = &\sum_{r \in \mathcal{D}^p}\sum_{w \in
      r}\sum_{w_t \in c_w}\log p(w_t|w) +\\
    &\sum_{r \in \mathcal{D}^p}\sum_{w \in
      r}\log p(y_w|w)
    \end{split}
  \end{equation}
  where $p(w_t|w)$ and $p(y_w|w)$ are defined in Eq.~\ref{eq:context} and Eq.~\ref{eq:senti} respectively. The likelihood of the reviews from $\mathcal{D}^q$, i.e $\mathcal{L}^q$, is defined similarly.

  For each word $w$ in the review $r$, in the E-step, the posterior probability of $z_w$ given $c_w$ and $y_w$ is:
  \begin{equation}
    \small
    \label{eq:post}
    \begin{split}
      &p(z_w=k|w, c_w, y_w) = \\
      &\frac{p(z_w=k)p(y_w|w, z_w=k)\prod \limits_{w_t \in
      c_w}p(w_t|w, z_w=k)}{\sum \limits_{k^{\prime}
        \in \{0, 1\}}p(z_w=k^{\prime})p(y_w|w,z_w=k^{\prime})\prod \limits_{w_t \in
        c_w}p(w_t|w, z_w=k^{\prime})}
  \end{split}
  \normalsize
    \end{equation}
  In the M-step, given the posterior distribution of $z_w$ in Eq.~\ref{eq:post}, the goal is to maxmize the following Q function:
  \begin{equation}
    \label{eq:q}
\begin{split}
\mathbf{Q} =& \sum \limits_{r\in\{\mathcal{D}^p, \mathcal{D}^q\}}\sum
\limits_{w \in r}\sum_{z_w} p(z_w|w, y_w, w_{t+j})\\
&\times
\log (p(z_w) p(c_w, y|z, w_t))\\
= &\sum \limits_{r\in\{\mathcal{D}^p, \mathcal{D}^q\}}\sum
\limits_{w \in r}\sum_{z_w} p(z_w|w, y_w, c_w) \\
& [\log p(z_w) + \log(y_w|z, w) + \\& \sum_{w_t \in c_w}\log p(w_t|z_w, w)] \\
\end{split}
\normalsize
\end{equation}

Using the Lagrange multiplier, we can obtain the update rule of $p(z_w)$, satisfying the normalization constraints that $\sum_{z_w \in {0, 1}} p(z_w)=1$ for each word $w$:
\begin{equation}
  \label{eq:z_w}
  p(z_w) = \frac{\sum_{r \in \{\mathcal{D}^p, \mathcal{D}^q\}}\sum_{w
      \in r}p(z_w|w, y_w, c_w)}{\sum_{r \in \{\mathcal{D}^p,
      \mathcal{D}^q\}} n(w, r)}
  \end{equation}
  where $n(w, r)$ is the number of occurrence of the word $w$ in the review $r$.

  To obtain $U_w^c$, $U_w^p$ and $U_w^q$,  we collect the related items in Eq.~\ref{eq:q} as follows:
  \begin{equation}
    \label{eq:qu}
    \begin{split}
    \mathbf{Q_U} = \sum \limits_{r\in\{\mathcal{D}^p, \mathcal{D}^q\}}\sum
    \limits_{w \in r}\sum_{z_w} p(z_w|w, y_w, w_{t+j}) \\
     [\log(y_w|z_w, w) + \sum_{w_t \in c_w}\log p(w_t|z_w,
     w)] \\
     \end{split}
   \end{equation}
   
  Note that computing the value $p(w_t|w, z_w)$ based on Eq.~\ref{eq:w_t} and Eq.~\ref{eq:w_t_z} is not feasible in practice, given that the computation cost is proportional to the size of $\Lambda$. However, similar to the skip-gram model, we can rely on negative sampling to address this issue. Therefore we estimate the probability of predicting the context word $p(w_{t}|w,z_w=1)$ as follows:
\begin{equation}
  \label{eq:sim}
\begin{split}
\log p(w_{t}|w, z_w=1) & \propto \log \sigma(U_{w}^c\cdot V_{w_t})\\
&+ \sum_{i=1}^{n}\mathbf{E}_{w_i\sim P}[\log \sigma(-U_{w}^c\cdot V_{w_i})]
\end{split}
\end{equation}
where $w_i$ is a negative instance which is sampled from the word distribution $P(.)$. Mikolov et al.~\shortcite{mikolov2013distributed} have investigated many choices for $P(w)$ and found that the best $P(w)$ is equal to the unigram distribution $\textit{Unigram}(w)$ raised to the $3/4rd$ power. We adopt the same setting. The probability $p(w_{t}|w,z_w=0)$ in Eq.~\ref{eq:w_t_z} can be approximated in a similar manner.
  
 After the substitution of $p(w_t|w, z_w)$, we use the Stochastic Gradient Descent method to maximize Eq.~\ref{eq:qu}, and obtain the update of $U_w^c$, $U_w^p$ and $U_w^q$.
 
 \subsection{More Discussions}
 In our model, for simplifying the inference algorithm and saving the computational cost, we assume that the target word $w_t$ in the context and the sentiment label $y_w$ of the word $w$ are conditionally independent.  Such technique has also been used in other popular models such as the bi-gram language model. Otherwise, we need to consider the term $p(w_t|w, y_w)$, which complicates the inference algorithm.

We define the formulation of the term $p(w_t|w, z)$ to be similar to the original skip-gram model instead of the CBOW model. The CBOW model averages the context words to predict the target word. The skip-gram model uses pairwise training examples which are much easier to integrate with sentiment information. 
 
 Note that our model can be easily extended to more than two domains. Similarly, we use a domain-specific vector for each word in each domain and each word is also associated with a domain-common vector. We just need to extend the probability distribution of $z_w$ from Bernoulli distribution to Multinomial distribution according to the number of domains.

 \section{Experiment}
 \begin{table*}[h]
   \small
  \begin{center}
\begin{tabular}{ |@{}c@{}|c@{~}|@{~}c@{~}||c@{~} |@{~}c@{~}||c@{~}|@{~}c@{~}||c@{~} |@{~}c@{~}||c@{~}|@{~}c@{~}||c@{~} |@{~}c@{~}|}
\hline
\multirow{2}{*}{}&\multicolumn{2}{c|}{{B} \& {D}} &\multicolumn{2}{c|}{{B} \& {E}}&\multicolumn{2}{c|}{{B} \& {K}}&\multicolumn{2}{c|}{{D} \& {E}}&\multicolumn{2}{c|}{{D} \& {K}}&\multicolumn{2}{c|}{{E} \& {K}}\\
\cline{2-13}
& \textit{Acc.} & F1 & \textit{Acc.} & F1 &\textit{Acc.} & F1 &\textit{Acc.} & F1 &\textit{Acc.} & F1 &\textit{Acc.} & F1 \\
\hline
{BOW}  & 0.680 & 0.653 & 0.738 & 0.720 & 0.734 & 0.725 & 0.705& 0.685& 0.706& 0.689& 0.739& 0.715\\
\hline
{EmbeddingP}  & 0.753 & 0.740  &0.752 &  0.745 & 0.742 & 0.741& 0.740 & 0.746& 0.707& 0.702& 0.761& 0.760\\
\hline
{EmbeddingQ} & 0.736 & 0.732  & 0.697  & 0.697 & 0.706& 0.701 & 0.762& 0.759 & 0.758& 0.759& 0.783& 0.780\\
\hline
{EmbeddingCat}  & 0.769 & 0.731  &0.768 &  0.763 & 0.763& 0.763& 0.787& 0.773& 0.770& 0.770& 0.807& 0.803\\
\hline
{EmbeddingAll}  & 0.769 & 0.759 & 0.765 & 0.740 & 0.775& 0.767& 0.783& 0.779& 0.779& 0.776& 0.819& 0.815\\
\hline
{Yang}  & 0.767 & 0.752 & 0.775 & 0.766 & 0.760 & 0.755 & 0.791 & 0.785 & 0.762& 0.760& 0.805 & 0.804 \\
\hline
{SSWE} & 0.783 & 0.772   & 0.791 & 0.780& {\bf 0.801} & 0.792 & 0.825 & 0.815& 0.795& 0.790& 0.835& 0.824\\
\hline
{$\text{DSE}_c$}  & 0.773 & 0.750  & 0.783 &  0.781 & 0.775 & 0.773 & 0.797 & 0.792 & 0.784 & 0.776 & 0.806& 0.800\\
\hline
{$\text{DSE}_w$}  & {\bf 0.794}$^{\dag\natural}$ & {\bf 0.793}$^{\dag\natural}$  &{\bf 0.806}$^{\dag\natural}$ &  {\bf 0.802}$^{\dag\natural}$ & 0.797$^{\dag}$& {\bf 0.793}$^{\dag}$& {\bf 0.843}$^{\dag\natural}$& {\bf 0.832}$^{\dag\natural}$& {\bf 0.829}$^{\dag\natural}$& {\bf 0.827}$^{\dag\natural}$& {\bf 0.856}$^{\dag\natural}$& {\bf 0.853}$^{\dag\natural}$\\
\hline
\end{tabular}
\normalsize
\caption{Results of review sentiment classification. The markers $^{\dag}$ and $^{\natural}$ refer to $p$-value $<$ 0.05 when comparing with Yang and SSWE respectively.}
\label{table:sentiment}
\end{center}
\end{table*}
 \subsection{Experimental Setup}
 We conducted experiments on the Amazon product reviews collected by Blitzer et al.~\shortcite{blitzer2007biographies}. We use four product categories: books (\textbf{B}), DVDs (\textbf{D}), electronic items (\textbf{E}), and kitchen appliances (\textbf{K}). A category corresponds to a domain. For each domain, there are 17,457 unlabeled reviews on average associated with rating scores from 1.0 to 5.0 for each domain. We use unlabeled reviews with rating score higher than 3.0 as positive reviews and unlabeled reviews with rating score lower than 3.0 as negative reviews for embedding learning. We first remove reviews whose length is less than 5 words. We also remove punctuations and the stop words. We also stem each word to its root form using Porter Stemmer~\cite{porter1980algorithm}. Note that this review data is used for embedding learning, and the learned embeddings are used as feature vectors of words to conduct the experiments in the later two subsections. 

Given the reviews from two domains, namely, $\mathcal{D}^p$ and $\mathcal{D}^q$, we compare our results with the following baselines and state-of-the-art methods:
\begin{description}
  \item[SSWE] The SSWE model\footnote{We use the implementation from \url{https://github.com/attardi/deepnl/wiki/Sentiment-Specific-Word-Embeddings}.} proposed by Tang et al.~\shortcite{tang2014learning} can learn sentiment-aware word embeddings from tweets. We employ this model on the combined reviews from $\mathcal{D}^p$ and $\mathcal{D}^q$ and then obtain the embeddings.
    
\item[Yang's Work]
    Yang et al.~\shortcite{yang2017simple} have proposed a method\footnote{We use the implementation from \url{http://statnlp.org/research/lr/}.} to learn domain-sensitive word embeddings. They choose pivot words and add a regularization item into the original skip-gram objective function enforcing that word representations of pivot words for the source and target domains should be similar. The method trains the embeddings of the source domain first and then fixes the learned embedding to train the embedding of the target domain. Therefore, the learned embedding of the target domain benefits from the source domain. We denote the method as Yang in short.
  
\item[EmbeddingAll] We learn word embeddings from the combined unlabeled review data of $\mathcal{D}^p$ and $\mathcal{D}^q$ using the skip-gram method~\cite{mikolov2013distributed}.
  
  \item[EmbeddingCat] We learn word embeddings from the unlabeled reviews of $\mathcal{D}^p$ and $\mathcal{D}^q$ respectively. To represent a word for review sentiment classification, we concatenate its learned word embeddings from the two domains.
    
  \item[EmbeddingP and EmbeddingQ] In EmbeddingP, we use the original skip-gram method~\cite{mikolov2013distributed} to learn word embeddings only from the unlabeled reviews of $\mathcal{D}^p$. Similarly, we only adopt the unlabeled reviews from $\mathcal{D}^q$ to learn embeddings in EmbeddingQ.
    
  \item[BOW] We use the traditional bag of words model to represent each review in the training data.
  \end{description}
  
  For our DSE model, we have two variants to represent each word. The first variant $\text{DSE}_c$ represents each word via concatenating the domain-common vector and the domain-specific vector. The second variant $\text{DSE}_w$ concatenates domain-common word embeddings and domain-specific word embeddings by considering the domain-commonality distribution $p(z_w)$. For individual review instances, the occurrences of w in $\mathcal{D}_p$ or $\mathcal{D}_q$ have been established. The representation of $w$ is specific to the occurrence of $w$ in $\mathcal{D}_p$ or $\mathcal{D}_q$. Specifically, each word $w$ can be represented as follows:
  \begin{equation}
    \begin{split}
      U_w =
      \begin{cases}
        \text{if
        } w \in \mathcal{D}^p \\
        \quad U_w^c \times p(z_w) \oplus U_w^p \times (1.0-p(z_w))  \\
         \text{if
        } w \in \mathcal{D}^q \\
        \quad U_w^c \times p(z_w) \oplus U_w^q \times (1.0-p(z_w))\\
        \end{cases}
        \end{split}
      \end{equation}
      where $\oplus$ denotes the concatenation operator.
      
For all word embedding methods, we set the dimension to 200. For the skip-gram based methods, we sample 5 negative instances and the size of the windows for each target word is 3. For our DSE model, the number of iterations for the whole reviews is 100 and the learning rate is set to 1.0.

\begin{table*}[t]
  \small
  \begin{center}
\begin{tabular}{ |@{~}c@{~}|c@{~}|@{~}c@{~}||c@{~} |@{~}c@{~}||c@{~}|@{~}c@{~}||c@{~} |@{~}c@{~}||c@{~}|@{~}c@{~}||c@{~} |@{~}c@{~}|}
\hline
\multirow{2}{*}{}&\multicolumn{2}{c|}{{B} \& {D}}
  &\multicolumn{2}{c|}{{ B}
                                                 \& {E}}&\multicolumn{2}{c|}{{ B}
                                                 \& {K}}&\multicolumn{2}{c|}{{ D}
                                                 \& {E}}&\multicolumn{2}{c|}{{ D}
                                                 \& {K}}&\multicolumn{2}{c|}{{ E}
                                                 \& {K}}\\
\cline{2-13}
& HL & MPQA & HL & MPQA &HL & MPQA
&HL & MPQA &HL & MPQA &HL & MPQA \\
\hline
{EmbeddingP}  & 0.740 & 0.733  & 0.742 &  0.734 & 0.747 & 0.735 &
                                                                  0.744& 0.701& 0.745& 0.709& 0.628 & 0.574\\
\hline
{EmbeddingQ} & 0.743 & 0.701  & 0.627  & 0.573 & 0.464 & 0.453 & 0.621 &
                                                                     0.577 &
                                                                           0.462 & 0.450 & 0.465 & 0.453\\
  \hline
{EmbeddingCat}  & 0.780 & 0.772  & 0.773 &  0.756 & 0.772 & 0.751 & 0.744 & 0.728 & 0.755 & 0.702 & 0.683 & 0.639\\
  \hline
    {EmbeddingAll}  & 0.777 & 0.769 & 0.773 & 0.730 & 0.762 & 0.760 & 0.712 & 0.707 & 0.749 & 0.724 & 0.670 & 0.658\\
  \hline
  {Yang}  & 0.780 & 0.775  & 0.789 &  0.762 & 0.781 & 0.770 & 0.762 & 0.736 & 0.756 &
                                                                      0.713&
                                                                            0.634 & 0.614\\
  \hline
  {SSWE} & {\bf 0.816} & {\bf 0.801}  & 0.831 & 0.817 & 0.822 & {\bf 0.808}
& {\bf 0.826} & 0.785 & 0.784 & 0.772 & 0.707 & 0.659\\
  \hline
  {$\text{DSE}$}  & 0.802 & 0.788  & {\bf 0.833} &  {\bf 0.828} & {\bf 0.832} & 0.799 & 0.804 & {\bf 0.797} & {\bf 0.796} & {\bf 0.786} & {\bf 0.725}& {\bf 0.683}\\
  \hline
\end{tabular}
\normalsize
\caption{Results of lexicon term sentiment classification.}
\label{table:lexicon_result}
\end{center}
\end{table*}

\subsection{Review Sentiment Classification}

 For the task of review sentiment classification, we use 1000 positive and 1000 negative sentiment reviews labeled by
 Blitzer et al.~\shortcite{blitzer2007biographies} for
 each domain to conduct experiments. We randomly select 800 positive and 800 negative labeled reviews from each domain as
 training data, and the remaining 200 positive and 200 negative labeled
 reviews as testing data. We use the SVM classifier~\cite{fan2008liblinear}
with linear kernel to train on the training reviews for each domain, with each review represented as the average vector of its word embeddings. 
 
 We use two metrics to evaluate the performance of sentiment
classification. One is the standard accuracy metric. The other one is
Macro-F1, which is the average of F1 scores for both positive and negative reviews.
 
We conduct multiple trials by selecting every possible two domains from books (\textbf{B}), DVDs
(\textbf{D}), electronic items (\textbf{E}) and kitchen appliances
(\textbf{K}). We use the average of the results of each two domains. The
experimental results are shown in Table~\ref{table:sentiment}.

From Table~\ref{table:sentiment}, we can see that
compared with other baseline methods, our $\text{DSE}_w$ model can achieve the
best performance of sentiment classification across most combinations of
the four domains. Our statistical t-tests for most of the combinations of domains show that the improvement of our $\text{DSE}_w$ model over Yang and SSWE is statistically significant respectively (p-value $< 0.05$) at 95\% confidence level. It shows that our method can capture the domain-commonality and sentiment information at the same time.

Even though
both of the SSWE model and our DSE model can learn sentiment-aware
word embeddings, our $\text{DSE}_w$ model can outperform
SSWE. It demonstrates that compared with general sentiment-aware embeddings, our learned domain-common and domain-specific word embeddings can capture semantic variations of words across multiple
domains.

Compared with the method of Yang which learns cross-domain embeddings, our $\text{DSE}_w$ model can
achieve better performance. It is because we exploit sentiment information to distinguish
domain-common and domain-specific words during the embedding learning
process. The sentiment information can also help the model
distinguish the words which have similar contexts but different
sentiments.

Compared with EmbeddingP and EmbeddingQ, the methods of EmbeddingAll
and EmbeddingCat can achieve better performance. The reason is that the
data augmentation from other domains helps sentiment classification in
the original domain. Our DSE model also benefits from such kind of
data augmentation with the use of reviews from $\mathcal{D}^p$ and $\mathcal{D}^q$.

We observe that our $\text{DSE}_w$ variant performs better than the variant of
$\text{DSE}_c$. Compared with $\text{DSE}_c$, our $\text{DSE}_w$ variant adds the item of
$p(z_w)$ as the weight to combine domain-common embeddings and
domain-specific embeddings. It shows that the domain-commonality
distribution in our DSE model, i.e $p(w_z)$, can effectively model the
domain-sensitive information of each word and help review sentiment classification.

\subsection{Lexicon Term Sentiment Classification}
To further evaluate the quality of the sentiment semantics of the learned word
embeddings, we also conduct lexicon term sentiment classification on two popular sentiment lexicons, namely
\textbf{HL}~\cite{hu2004mining} and
\textbf{MPQA}~\cite{wilson2005recognizing}. The words with neutral
sentiment and phrases are removed. The statistics of \textbf{HL}
and \textbf{MPQA} are shown in Table~\ref{table:lexicon}. 

\begin{table}[h]
\begin{center}
  \begin{tabular}{|l|rl|c|}
\hline Lexicon & Positive & Negative & Total \\ \hline
HL  & 1,331 & 2,647 & 3,978 \\
MPQA & 1,932 & 2,817 & 3,075 \\
\hline
\end{tabular}
\end{center}
\caption{Statistics of the sentiment lexicons.}
      \label{table:lexicon}
\end{table}

We conduct multiple trials by selecting every possible two domains from books (\textbf{B}), DVDs
(\textbf{D}), electronics items (\textbf{E}) and kitchen appliances
(\textbf{K}). For each trial, we learn the word embeddings. For
our DSE model, we only use the domain-common part to represent each
word because the lexicons are usually not associated with a
particular domain. For each lexicon, we select 80\% to train the SVM
classifier with linear kernel and the remaining 20\% to test
the performance. The learned embedding is treated as the feature
vector for the lexicon term. We conduct 5-fold cross validation on all the
lexicons. The evaluation metric is Macro-F1 of positive and
negative lexicons.

Table~\ref{table:lexicon_result} shows the experimental results of lexicon term sentiment classification. Our DSE method can achieve
competitive performance among all the methods. Compared with SSWE, our
DSE is still competitive because both of them consider the sentiment
information in the embeddings. Our DSE model outperforms other
methods which do not consider sentiments such as Yang,
EmbeddingCat and EmbeddingAll. Note that the advantage of domain-sensitive
embeddings would be insufficient for this task because the sentiment
lexicons are not domain-specific.

\section{Case Study}
\begin{table*}[h]
  \begin{center}
  \begin{small}
    \begin{tabular}{|c|c|c|c|}
      \hline
      Term & Domain & $p(z=1)$ & Sample Reviews \\
      \hline
      \multirow{6}{*}{``lightweight''} & B \& D & 0.999 & \multirow{6}{300pt}{\small - I find Seth Godin's books incredibly {\bf lightweight}. There is really nothing of any substance here.({\tt B}) \\ - I love the fact that it's small and {\bf lightweight} and fits into a tiny pocket on my camera case so I never lose track of it.({\tt E})\\ - These are not "{\bf lightweight}" actors. ({\tt D})\\ - This vacuum does a pretty good job.  It is {\bf lightweight} and easy to use.({\tt K})} \\ 
      \cline{2-3}
           & B \& E & 0.404 & \\
      \cline{2-3}
           & B \& K & 0.241 & \\
      \cline{2-3}
           & D \& E & 0.380 & \\
      \cline{2-3}
      & D \& K & 0.013 &\\
      \cline{2-3}
           & E \& K & 0.696 & \\
      \hline

            \multirow{6}{*}{``die''} & \multirow{2}{*}{B \& E} & \multirow{2}{*}{0.435} & \multirow{6}{300pt}{\small - I'm glad Brando lived long enough to get old and fat, and that he didn't {\bf die} tragically young like Marilyn, JFK, or Jimi Hendrix.({\tt B})\\ - Like many others here, my CD-changer {\bf died} after a couple of weeks and it wouldn't read any CD.({\tt E}) \\ - I had this toaster for under 3 years when I came home one day and it smoked and {\bf died}. ({\tt K})
      } \\
           &  &  & \\
      \cline{2-3}
           & \multirow{2}{*}{B \& K} & \multirow{2}{*}{0.492} & \\
           &  & & \\
      \cline{2-3}
           &  \multirow{2}{*}{E \& K} & \multirow{2}{*}{0.712}  & \\
           &  & & \\
      \hline
      
      \multirow{5}{*}{``mysterious''} & & &  \multirow{5}{300pt}{\small - This novel really does cover the gamut: stunning twists, genuine love, beautiful settings, desire for riches, {\bf mysterious} murders, detective investigations, false accusations, and self vindication.({\tt B}) \\
      - Caller ID functionality for Vonage {\bf mysteriously} stopped working even though this phone's REN is rated at 0.1b. ({\tt E})
      } \\
           &  &  & \\
           &  {B \& E} & {0.297}  & \\
           &  &  & \\
           &  &  & \\
      \hline
      \multirow{6}{*}{``great''} & B \& D & 0.760 & \multirow{6}{300pt}{\small - This is a {\bf great} book for anyone learning how to handle dogs.({\tt B}) \\ - This is a {\bf great} product, and you can get it, along with any other products on Amazon up to \$500 Free!({\tt E}) \\ - I grew up with drag racing in the 50s, 60s \& 70s and this film gives a {\bf great} view of what it was like.({\tt D}) \\ - This is a {\bf great} mixer its a little loud but worth it for the power you get.({\tt K})
      } \\
      \cline{2-3}
           & B \& E & 0.603 & \\
      \cline{2-3}
           & B \& K & 0.628 & \\
       \cline{2-3}
           & D \& E & 0.804 & \\
      \cline{2-3}
           & D \& K & 0.582 & \\
      \cline{2-3}
           & E \& K & 0.805 & \\
      \hline
    \end{tabular}
    \end{small}
    \caption{Learned domain-commonality for some words. $p(z=1)$ denotes the probability that the word is domain-common. The letter in parentheses indicates the domain of the review.}
    \label{table:prob}
    \end{center}
  \end{table*}
Table~\ref{table:prob} shows the probabilities of ``lightweight'', ``die'', ``mysterious'', and ``great'' to be domain-common for different domain combinations. For ``lightweight'', its domain-common probability for the books domain and the DVDs domain (``B \& D'') is quite high, i.e. $p(z=1)=0.999$, and the review examples in the last column show that the word ``lightweight'' expresses the meaning of lacking depth of content in books or movies. Note that most reviews of DVDs are about movies. In the electronics domain and the kitchen appliances domain (``E \& K''),  ``lightweight'' means light material or weighing less than average, thus the domain-common probability for these two domains is also high, 0.696. In contrast, for the other combinations, the probability of ``lightweight'' to be domain-common is much smaller, which indicates that the meaning of ``lightweight'' varies. Similarly, ``die'' in the domains of electronics and kitchen appliances  (``E \& K'') means that something does not work any more, thus, we have $p(z=1)=0.712$. While for the books domain, it conveys meaning that somebody passed away in some stories. The word ``mysterious'' conveys a positive sentiment in the books domain, indicating how wonderful a story is, but it conveys a negative sentiment in the electronics domain typically describing that a product breaks down unpredictably. Thus, its domain-common probability is small. The last example is the word ``great'', and it usually has positive sentiment in all domains, thus has large values of $p(z=1)$ for all domain combinations.

\section{Conclusions}
We propose a new method of learning domain-sensitive and
sentiment-aware word embeddings. Compared with existing sentiment-aware
embeddings, our model can distinguish domain-common and
domain-specific words with the consideration of varied semantics
across multiple domains. Compared with existing domain-sensitive methods,
our model detects domain-common words according to not only similar context
words but also sentiment information. Moreover, our learned embeddings
considering sentiment information can distinguish words with similar
syntactic context but opposite sentiment polarity. We have conducted
experiments on two downstream sentiment classification tasks, namely review sentiment
classification and lexicon term sentiment classification. The
experimental results demonstrate the advantages of our approach.

\bibliography{acl2018}
\bibliographystyle{acl_natbib}
\end{document}